\documentclass[10pt,twocolumn,letterpaper]{article}
\pdfoutput=1

\usepackage{iccv}
\usepackage{times}
\usepackage{epsfig}
\usepackage{graphicx}
\usepackage{amsmath}
\usepackage{amssymb}
\usepackage{multirow}
\usepackage{array}
\usepackage{tabularx}
\usepackage{boldline}

\newcolumntype{Y}{>{\centering\arraybackslash}X}

\usepackage[pagebackref=true,breaklinks=true,letterpaper=true,colorlinks,bookmarks=false]{hyperref}

\iccvfinalcopy 

\def\iccvPaperID{2102} 

\begin{document}

\title{Learning Anchored Unsigned Distance Functions with Gradient Direction Alignment for Single-view Garment Reconstruction}

\author{Fang Zhao$^1$ \quad \quad \quad Wenhao Wang$^2$ \quad \quad \quad Shengcai Liao$^{1}$\thanks{Corresponding author.} \quad \quad \quad Ling Shao$^{1,3}$ \\
$^1$Inception Institute of Artificial Intelligence \ \ \  $^2$ReLER, University of Technology Sydney \\
$^3$Mohamed bin Zayed University of Artificial Intelligence
}


\maketitle

\begin{abstract}
	While single-view 3D reconstruction has made significant progress benefiting from deep shape representations in recent years, garment reconstruction is still not solved well due to open surfaces, diverse topologies and complex geometric details. In this paper, we propose a novel learnable Anchored Unsigned Distance Function (AnchorUDF) representation for 3D garment reconstruction from a single image. AnchorUDF represents 3D shapes by predicting unsigned distance fields (UDFs) to enable open garment surface modeling at arbitrary resolution. To capture diverse garment topologies, AnchorUDF not only computes pixel-aligned local image features of query points, but also leverages a set of anchor points located around the surface to enrich 3D position features for query points, which provides stronger 3D space context for the distance function. Furthermore, in order to obtain more accurate point projection direction at inference, we explicitly align the spatial gradient direction of AnchorUDF with the ground-truth direction to the surface during training. Extensive experiments on two public 3D garment datasets, i.e., MGN and Deep Fashion3D, demonstrate that AnchorUDF achieves the state-of-the-art performance on single-view garment reconstruction. Code is available at \url{https://github.com/zhaofang0627/AnchorUDF}.
\end{abstract}

\section{Introduction}

 \begin{figure*}[t]
	\centering
	\includegraphics[width=17.0cm]{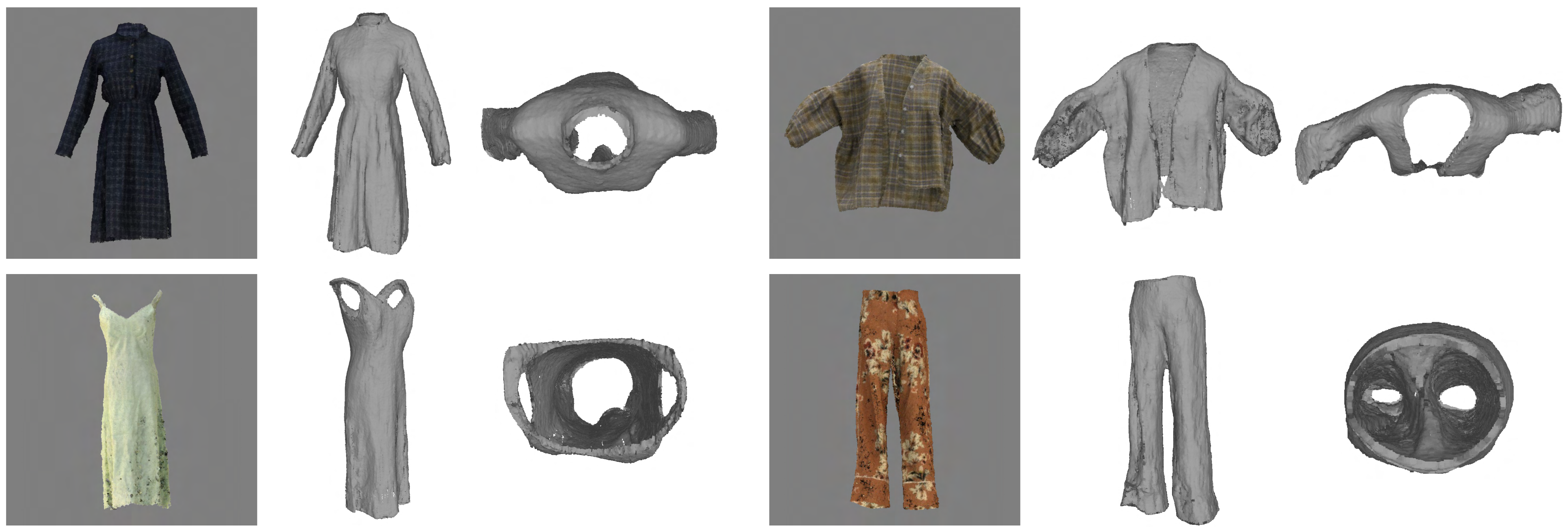}
	\caption{Single-view garment reconstruction using our method, which handles non-closed garment surfaces, captures multiple topologies of different garment categories and retains fine-scale geometric details.}\label{fig_1}
\end{figure*}

3D garment reconstruction has a wide range of applications in clothed human digitization, virtual try-on, online shopping and so on. Recently, image based 3D reconstruction has made significant progress benefiting from shape representation learning with deep neural networks~\cite{choy20163d,fan2017point,groueix2018,wang2018pixel2mesh,mescheder2019occupancy}. Compared to voxels, points and meshes, implicit functions, which define a surface as a level set of a function, can represent 3D surfaces at arbitrary resolution and produce fine-scale detailed surfaces in a memory-efficient way, and have been successfully applied to single-view human reconstruction~\cite{saito2019pifu,saito2020pifuhd}. However, recovering 3D garment shape from a single image is still a challenging task because garments have open surfaces and diverse topologies in addition to complex geometric details like clothed humans.

Existing methods usually use parametric models to provide shape priors of garments~\cite{bhatnagar2019multi,jiang2020bcnet,zhu2020deep}. BCNet~\cite{jiang2020bcnet} introduces a layered garment representation on top of the SMPL body model~\cite{loper2015smpl} and a generic skinning weights generating network to improve the expression ability of the garment model. Deep Fashion3D~\cite{zhu2020deep} proposes adaptable template meshes to fit garment shapes and incorporates implicit representations to refine surface details. These methods rely on pre-defined category-specific templates and have poor scalability to new garment categories.

In this paper, we propose Anchored Unsigned Distance Function (AnchorUDF), a learnable unsigned distance function that enriches 3D position features of query points by anchor points around 3D surfaces. With AnchorUDF, we establish a unified shape learning framework to reconstruct 3D garment from a single image. As shown in Fig.~\ref{fig_1}, our method can handle non-closed garment surfaces, capture multiple topologies of different garment categories while retaining fine-scale geometric details.

Specifically, for a 3D query point, like PIFu~\cite{saito2019pifu}, we first compute its pixel-aligned local image features. However, instead of using an absolute depth value to encode the 3D position of the query point, we leverage a set of anchor points representing 3D shape profile to enrich its position feature to make the distance function better sense the garment topologies. To obtain a small number of anchor points which can adequately cover the surface, we cluster points sampled from the surface with k-means and use clustering centers as targets to learn a regression network on top of the backbone to predict anchor points. Note that compared to the full shape, these few anchor points (typically a few hundred) are easier to estimate. A 3D convolutional network is further employed to compute 3D feature tensors from anchor points and a 3D position feature vector for the query point can be extracted via trilinear interpolation with its 3D coordinates, which encodes the relative position relationship between query and anchor points and provides stronger 3D space context information compared to the absolute depth value. Different from signed distance functions~\cite{park2019deepsdf}, we need to compute the gradient field of unsigned ones at inference to project the query point onto the surface along the negative gradient direction. Thus, in order to obtain more accurate estimation for the projection direction, we explicitly constrain the spatial gradient of our AnchorUDF during training to align its direction with the ground-truth direction to the surface.

Our contributions can be summarized as follows: 1) We propose a unified shape learning framework for single-view garment reconstruction; 2) We introduce AnchorUDF, a learnable unsigned distance function with anchored 3D position features; 3) We propose to learn the unsigned distance function with gradient direction alignment to more accurately project query points at inference.

 \begin{figure*}[t]
	\centering
	\includegraphics[width=17.0cm]{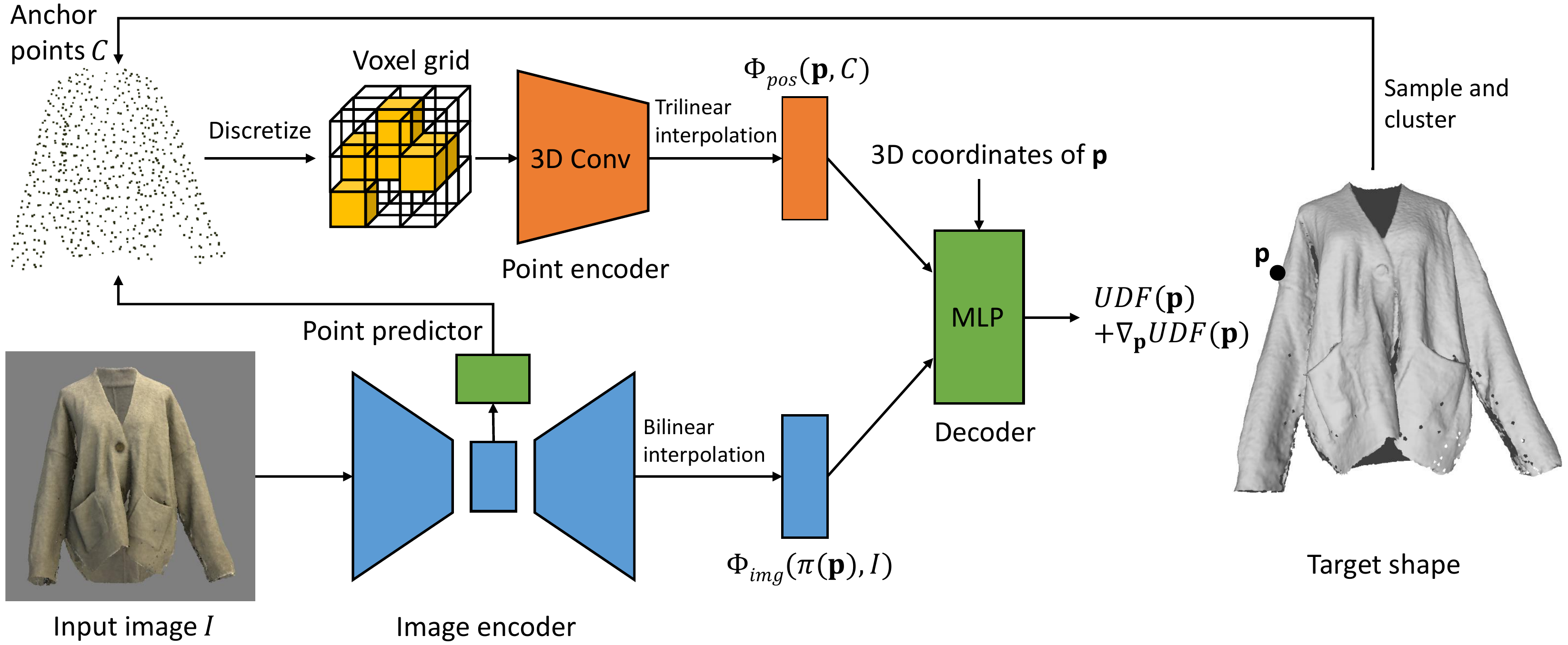}
	\caption{Framework of our proposed method. AnchorUDF predicts the unsigned distance field (UDF) of surface by using both local image features ${\Phi _{img}}(\pi ({\mathbf{p}}),I)$ and 3D position features $\Phi_{pos} ({\mathbf{p}},\mathcal{C})$ based on a set of anchor points. During training, we explicitly constrain the spatial gradient of AnchorUDF to align its direction with the ground-truth direction to the surface.}\label{fig_2}
	\vspace{-8pt}
\end{figure*}

\section{Related Work}

\noindent\textbf{Single-view garment/clothed body reconstruction.} Current work can generally be divided into template-based and template-free methods. For the former one, parametric 3D models~\cite{anguelov2005scape,loper2015smpl,pavlakos2019expressive} are used to provide strong priors for constraining the solution space of shape estimation. For better geometric detail representation, a high-frequency displacement is usually computed on the basis of the mesh model~\cite{danvevrek2017deepgarment,alldieck2018detailed,yu2019simulcap,alldieck2019learning,zhu2019detailed,zhu2020deep,jiang2020bcnet}. DeepWrinkles~\cite{lahner2018deepwrinkles} jointly represents global shape deformation and surface details by adding fine clothing wrinkles onto normal maps of a coarse garment mesh. Tex2shape~\cite{alldieck2019tex2shape} regards shape regression as an aligned image-to-image translation problem and estimates detailed normal and vector displacement maps from a partial texture, which can be applied to a body model to add detail and clothing. MGN~\cite{bhatnagar2019multi} predicts PCA coefficients of garment parametric models and a displacement field on top of PCA for clothing details. However, these template-based methods usually are limited by topologies of pre-defined garment template meshes and are hard to handle out-of-scope deformations.

Some template-free methods which do not use parametric models have been proposed to directly regress 3D shapes for modeling more complex garment topologies~\cite{varol2018bodynet,agudo2018scalable,saito2019pifu}. Bodynet~\cite{varol2018bodynet} infers volumetric body shape from a single image by an end-to-end trainable network with intermediate supervision of pose and body part segmentation. DeepHuman~\cite{zheng2019deephuman} introduces an image-guided volume-to-volume translation CNN by taking a semantic body volume as an additional input. These methods based on voxel representations require high memory and often fail to obtain fine-scale shape details. Based on implicit function representations, PIFu~\cite{saito2019pifu} proposes a memory-efficient deep learning framework for clothed body reconstruction, which locally aligns 2D image pixels with the global context of 3D objects to retain more shape details. More recently, PIFuHD~\cite{saito2020pifuhd} formulates a multi-level architecture based on PIFu, where a coarse level focuses on holistic reasoning and a fine level estimates highly detailed geometry. Geo-PIFu~\cite{he2020geo} extends PIFu to learn latent voxel features using a structure-aware 3D U-Net for less shape distortion and sharper surface details. There also exist some methods~\cite{bhatnagar2020combining,huang2020arch} that combine implicit functions and parametric models to obtain both detailed and controllable 3D body reconstructions.

\noindent\textbf{Differences from related reconstruction methods.} PIFuHD~\cite{saito2020pifuhd} uses larger input image resolution and extra normal maps for higher-fidelity reconstruction. Our method can also incorporate the HD module to further improve the reconstruction performance as demonstrated in our experiments. ARCH~\cite{huang2020arch} requires a body mesh model to obtain body landmarks, which is hard to extend to garments which have multiple topologies. Geo-PIFu~\cite{he2020geo} uses an extra 3D backbone network to regress dense 3D volumes, which is computationally intensive for both training and testing. In contrast, our method does not rely on external parametric models, and only adds a lightweight head net on the top of backbone for anchor point prediction and a small 3D convolutional network for point feature extraction. Besides, the most important point is that the aforementioned methods are only able to generate closed surfaces and thus cannot handle open boundaries of garments.

\noindent\textbf{Implicit surface representation.} Compared to voxel~\cite{wu20153d,wu2016learning,choy20163d,gadelha20173d}, point~\cite{fan2017point,achlioptas2018learning} and mesh~\cite{kanazawa2018learning,ranjan2018generating} representations, implicit functions allow to represent 3D surfaces at infinite resolution without excessive memory cost~\cite{mescheder2019occupancy,park2019deepsdf,mullen2010signing,atzmon2020sal,chibane2020neural}. OccNet~\cite{mescheder2019occupancy} proposes to encode the 3D surface as the decision boundary of a deep neural network classifier. DeepSDF~\cite{park2019deepsdf} introduces a learned continuous signed distance function for a class of shapes, which implicitly represents a shape’s boundary as the zero level-set of the learned function. IF-Nets~\cite{chibane2020implicit} learn implicit functions by extracting and classifying deep features extracted from a 3D grid of multi-scale features at a continuous query point. SAL~\cite{atzmon2020sal} defines a family of loss functions for sign agnostic learning with raw geometric data. IGR~\cite{gropp2020implicit} encourages the neural network to vanish on input points and has a unit norm gradient to learn smooth and natural implicit surfaces. To enable implicit functions to model open surfaces, NDF~\cite{chibane2020neural} proposes to predict unsigned distance fields (UDFs) to represent open surfaces for point cloud completion. However, it is nontrivial to learn UDFs for image based garment reconstruction due to lack of 3D space information. SALD~\cite{atzmon2020sald} includes derivatives during sign agnostic learning, which leads to a lower sample complexity and better fitting. Different from its motivation and loss form, we aim to optimize the gradient direction of UDF to obtain more accurate point projection direction at inference.

\section{Method}

Given a single image, we aim to recover the 3D garment shape with open surfaces and fine-scale surface details. In this section, we propose Anchored Unsigned Distance Function (AnchorUDF) which predicts the unsigned distance field (UDF) of surface with anchored 3D position features. To project query points onto the surface along more accurate direction at inference, we also explicitly align the spatial gradient direction of AnchorUDF with the ground-truth direction to the surface. The framework of our proposed method is illustrated in Fig.~\ref{fig_2}.

\subsection{Unsigned Distance Fields}

To represent shapes with open surfaces, we adopt unsigned distance fields (UDFs)~\cite{chibane2020neural} which assign a non-negative scalar value $s$ to a spatial point $\mathbf{p}$:
\begin{equation}\label{eq_udf}
UDF({\mathbf{p}}) = s:{\mathbf{p}} \in {\mathbb{R}^3},s \in \mathbb{R}_0^ + ,
\end{equation}
where $s$ represents the unsigned distance from $\mathbf{p}$ to the closest surface and the shape surface is implicitly represented by the zero level-set $UDF(.) = 0$. In contrast to signed distance fields (SDFs)~\cite{park2019deepsdf} or occupancies~\cite{mescheder2019occupancy} which divide the 3D space into inside and outside the surface, UDFs allow to naturally represent open surfaces. We can project $\mathbf{p}$ onto the surface by moving $\mathbf{p}$ along the negative gradient direction of UDF:
\begin{equation}\label{eq_project}
{\mathbf{q}}: = {\mathbf{p}} - UDF({\mathbf{p}}) \cdot {\nabla _{\mathbf{p}}}UDF({\mathbf{p}}).
\end{equation}
Dense point clouds are able to be easily computed from the implicit surfaces using fast gradient evaluation for UDFs~\cite{chibane2020neural} and naive classical algorithms for meshing~\cite{bernardini1999ball} can be used to generate the corresponding meshes.

\subsection{Anchored Unsigned Distance Functions}

It is worth noting that regressing a continuous distance field is much more difficult than classifying a binary occupancy value. Moreover, different from SDFs, UDFs need gradient direction computation at inference. Besides a point itself, the prediction accuracy for the neighborhood of the point is also required to ensure the correct gradient direction. Thus, discriminative features are more critical for learning a good UDF.

To this end, our AnchorUDF employs both local image features and 3D position features of query points to predict UDF values. The key idea is to leverage a set of anchor points located around 3D surfaces to compute relative position features for query points, which makes the distance function better fit the garment topologies.

Given an input image $I$, a fully convolutional image encoder is first used to compute a feature map of $I$. For a 3D query point $\mathbf{p}$, we project it onto its corresponding position on the image plane to compute its local image features from the feature map by bilinear interpolation at the projected pixel coordinates: ${\Phi _{img}}(\pi ({\mathbf{p}}),I)$, where $\pi$ represents the (weak) perspective camera projection.

In order to compute more discriminative 3D position features for the query point $\mathbf{p}$, a set of anchor points $\mathcal{C}$ located around the surface is introduced to encode the 3D position of $\mathbf{p}$. We then discretize the anchor points $\mathcal{C}$ to a voxel grid and feed the grid into a point encoder consisting of a series of 3D convolutional layers to produce a 3D feature tensor of $\mathcal{C}$. By applying trilinear interpolation to the feature tensor according to the 3D coordinates of $\mathbf{p}$, we extract the feature vector $\Phi_{pos} ({\mathbf{p}},\mathcal{C})$ as the position features of the query point. Compared to using an absolute depth value to provide the 3D position information~\cite{saito2019pifu}, our position features reflect relative position relationship between query and anchor points and provides stronger 3D space context to make UDF better capture multiple topologies.

To obtain a set of anchor points which can adequately cover the surface only using a small number of points, we cluster points sampled from the ground-truth surface with k-means and use clustering centers $\tilde {\mathcal{C}}$ as targets to learn a regression network on top of the backbone to predict the anchor points. Here the anchor point prediction and the shape reconstruction are learned jointly, which can be seen as multi-task learning since they share the backbone. We argue that compared to the full shape, these few anchor points (typically a few hundred) are easier to estimate. Therefore, our method can also be regarded as a coarse-to-fine strategy, i.e., a shape profile is first predicted, then further refined to produce the detailed surface.

At last, we concatenate the local image features, the 3D position features and the 3D coordinates of the query point as the input of a decoder to predict UDF values and our AnchorUDF is formulated as:
\begin{equation}\label{eq_feat}
{f_I}({\mathbf{p}},\mathcal{C};{\mathbf{w}}) = {f_{dec}}([{\Phi _{img}}(\pi ({\mathbf{p}}),I),{\Phi _{pos}}({\mathbf{p}},\mathcal{C}),{\mathbf{p}}]),
\end{equation}
where $\mathbf{w}$ denotes network parameters and the decoder $f_{dec}$ is parameterized by multi-layer perceptrons (MLP) with ReLU in its last layer.

 \begin{figure}[t]
	\centering
	\includegraphics[width=6.0cm]{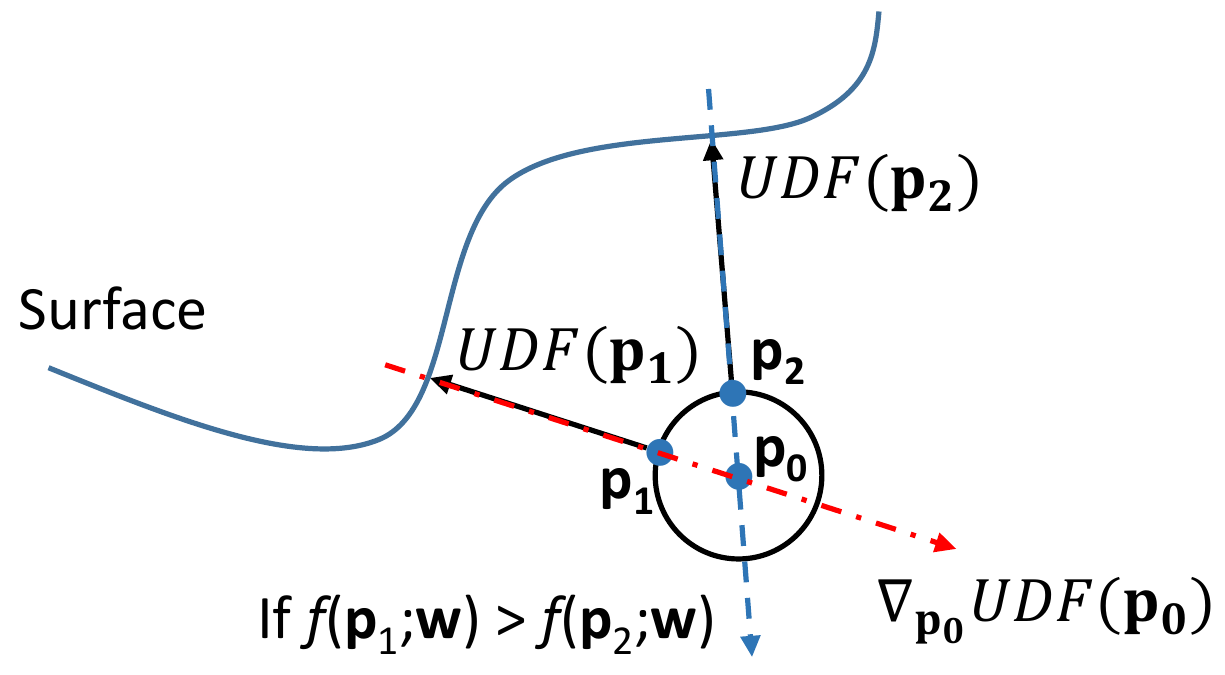}
	\caption{Motivation of gradient direction alignment. ${\mathbf{p}}_1$ and ${\mathbf{p}}_2$ are two points on the neighborhood of ${\mathbf{p}}_0$. When $UDF({\mathbf{p}}_1)$ and $UDF({\mathbf{p}}_2)$ are close, even if the distance loss is small, it is also possible that $f({{\mathbf{p}}_1};{\mathbf{w}}) > f({{\mathbf{p}}_2};{\mathbf{w}})$. In this case, the predicted gradient direction at ${\mathbf{p}}_0$ (the blue arrow) will be significantly different from the direction of ${\nabla _{{{\mathbf{p}}_0}}}UDF({{\mathbf{p}}_0})$ (the red arrow).
	}\label{fig_grad}
	\vspace{-8pt}
\end{figure}

\subsection{Learning with Gradient Direction Alignment}

To learn our AnchorUDF, for an input image $I$, we generate training examples $\mathcal{P}_I$ by sampling points near its corresponding surface and computing their ground-truth UDF values $UDF({\mathbf{p}})$. Then, we minimize the L1 loss between the predicted and ground-truth UDF values on $\mathcal{P}_I$ by updating parameters $\mathbf{w}$ of ${f_I}({\mathbf{p}},\mathcal{C};{\mathbf{w}})$:
\begin{equation}\label{loss_df}
{L_{UDF}} = \sum\limits_{{\mathbf{p}} \in \mathcal{P}_I} {|\min ({f_I}({\mathbf{p}},\mathcal{C};{\mathbf{w}}),\delta ) - \min ({UDF({\mathbf{p}})},\delta )|}.
\end{equation}
Similar to~\cite{chibane2020neural,park2019deepsdf}, a small value $\delta$ is used to clamp the maximal regressed distance to concentrate the model capacity on details in the vicinity of the surface.

In order to learn the anchor point predictor, we consider a loss defined by the Chamfer distance~\cite{fan2017point} between the predicted anchor points $\mathcal{C}$ and the targets $\tilde {\mathcal{C}}$ because the set of anchor points is unordered:
\begin{equation}\label{loss_cd}
{L_{AP}} = \sum\limits_{{\mathbf{c}} \in {\mathcal{C}}} {\mathop {\min }\limits_{{{\mathbf{\tilde c}}} \in \tilde {\mathcal{C}}} ||{\mathbf{c}} - {{\mathbf{\tilde c}}}||_2^2}  + \sum\limits_{{{\mathbf{\tilde c}}} \in \tilde {\mathcal{C}}} {\mathop {\min }\limits_{{\mathbf{c}} \in {\mathcal{C}}} ||{\mathbf{c}} - {{\mathbf{\tilde c}}}||_2^2} .
\end{equation}

The standard back-propagation through AnchorUDF can be used to compute the spatial gradients of the learned distance field to project points by Eq.~\eqref{eq_project}.

However, only optimizing the point-wise distance loss~\eqref{loss_df} cannot guarantee a good estimation for the true gradient directions of UDF. As illustrated in Fig.~\ref{fig_grad} (for the sake of brevity, here $I$ and $\mathcal{C}$ in ${f_I}({\mathbf{p}},\mathcal{C};{\mathbf{w}})$ are omitted), consider two points ${\mathbf{p}}_1$ and ${\mathbf{p}}_2$ on the neighborhood of a point ${\mathbf{p}}_0$, when the ground truth UDF values of ${\mathbf{p}}_1$ and ${\mathbf{p}}_2$ are close, because the distance loss does not penalize the relationship between points, even if the loss is small, it is also possible that $f({{\mathbf{p}}_1};{\mathbf{w}}) > f({{\mathbf{p}}_2};{\mathbf{w}})$. In this case, the predicted gradient direction at ${\mathbf{p}}_0$ will be significantly different from the ground truth one. Therefore, during training we explicitly constrain the spatial gradient of AnchorUDF to align its direction with the true gradient direction by the following loss:
\begin{equation}\label{loss_grad}
{L_{GDA}} = \sum\limits_{{\mathbf{p}} \in \mathcal{P}} {1 - \cos ( {\nabla _{\mathbf{p}}}f({\mathbf{p}};{\mathbf{w}}),{\nabla _{\mathbf{p}}}UDF({\mathbf{p}}))}.
\end{equation}
In practice, the direction of ${\nabla _{\mathbf{p}}}UDF({\mathbf{p}})$ can be computed by $({\mathbf{p}}-{\mathbf{q}})/||{\mathbf{p}} - {\mathbf{q}}||_2$, where ${\mathbf{q}}$ is the point closest to ${\mathbf{p}}$ on the surface.

Finally, the overall objective function is
\begin{equation}\label{loss_total}
L = {L_{UDF}} + {\lambda _1}{L_{AP}} + {\lambda _2}{L_{GDA}},
\end{equation}
where $\lambda _1$ and $\lambda _2$ are loss weights.

 \begin{figure}[t]
	\centering
	\includegraphics[width=8.0cm]{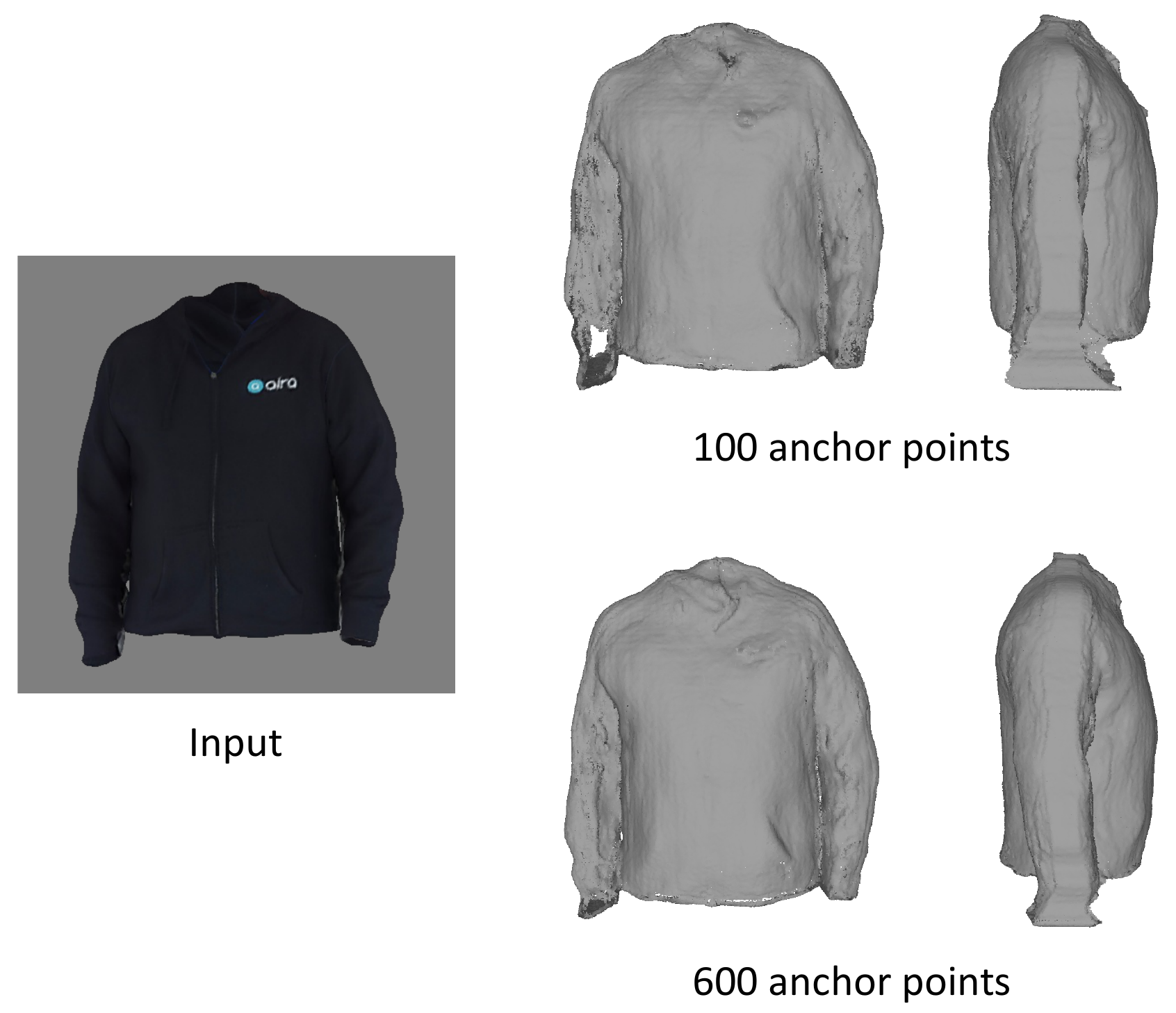}
	\caption{Visual comparison of using different numbers of anchor points. More anchor points can better project query points to the surfaces}\label{fig_ap_num}
	\vspace{-8pt}
\end{figure}

\section{Experiments}

We evaluate the proposed AnchorUDF on two 3D garment datasets, i.e, MGN~\cite{bhatnagar2019multi} and Deep Fashion3D~\cite{zhu2020deep}. We adopt the Chamfer distance and the average point-to-surface Euclidean distance (P2S) to measure the quality of shape reconstruction. We also compare AnchorUDF against other single-view 3D reconstruction methods with different shape representations~\cite{wang2018pixel2mesh,saito2019pifu,jiang2020bcnet,saito2020pifuhd}.

 \begin{figure}[t]
	\centering
	\includegraphics[width=8.3cm]{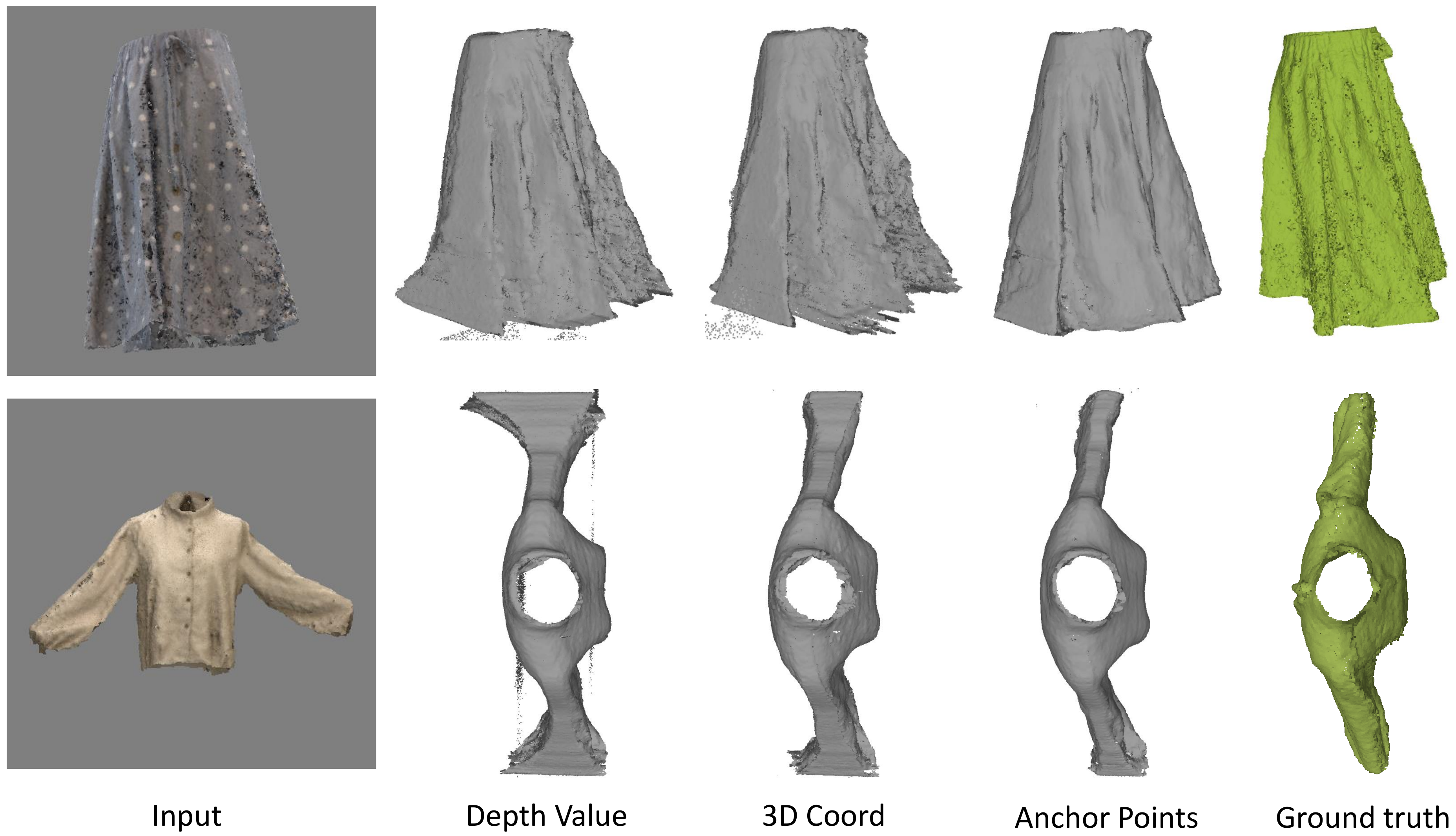}
	\caption{Visual comparison of using different 3D position features. Our anchor point based position features can better recover surfaces, especially along the depth direction of input images}\label{fig_pos_feat}
\end{figure}

\noindent\textbf{Datasets.} MGN~\cite{bhatnagar2019multi} contains 5 garment categories and 154 textured garments models. 134 garments models are randomly selected as the training set and the remaining 20 models form the test set. Following PIFu, we render images with 360 degrees in yaw axis for each garment model and obtain 48,240 images for training. Deep Fashion3D~\cite{zhu2020deep} is a large-scale collection of 3D garment models with diverse shapes and poses. It consists of 2075 garment models covering 10 different categories and 598 instances. We use 1880 garment models for training and 195 models for testing, where the training and test sets have disjoint instances. Because the dataset does not release the corresponded multi-view real images, we only use rendered images as the training inputs. For each model, we render images by sampling 18 front viewpoints, resulting in 33,840 training images. Similar to~\cite{zhu2020deep}, here we focus on front-view reconstruction. 

\begin{table}[t]
	\centering
	\caption{Chamfer and P2S errors ($ \times {10^{ - 3}}$) of using different numbers of anchor points (AP) on MGN dataset.}
	\label{tab_num_ap}
	\begin{tabularx}{0.47\textwidth}{lYYYY@{}}
		\hlineB{2.5}
		Num. AP & 100  & 300 & 600 & 900 \\
		\hline
		Chamfer &  0.731 & 0.716  & \textbf{0.696} & 0.758  \\
		P2S &  0.987  &  0.967  & \textbf{0.899}  & 1.014  \\
		\hlineB{2.5}
	\end{tabularx}
\end{table}

\begin{table}[t]
	\centering
	\caption{Chamfer and P2S errors ($ \times {10^{ - 3}}$) using different 3D position features on MGN and Deep Fashion3D datasets.}
	\label{tab_results_ap}
	\begin{tabularx}{0.47\textwidth}{lYYYY@{}}
		\hlineB{2.5}
		\multirow{2}{*}{Methods} & \multicolumn{2}{c}{MGN} & \multicolumn{2}{c}{Deep Fashion3D} \\
		\cline{2-5}
		& Chamfer & P2S & Chamfer & P2S \\
		\hline
		Depth Value & 1.063  & 1.974 & 1.411 & 2.440 \\
		3D Coord & 0.757 & 1.060 & 1.062 & 1.748 \\
		Anchor Points & \textbf{0.696} & \textbf{0.899} & \textbf{0.712} & \textbf{0.932} \\
		\hlineB{2.5}
	\end{tabularx}
\end{table}

\noindent\textbf{Implementation Details.} We adopt a stacked hourglass~\cite{newell2016stacked} network with 5 stacks as our backbone to encode images. For anchor point prediction, we add a 4-layer network on top of the backbone, which consists of 3 convolutional (Conv) layers and 1 fully connected (FC) layer. To extract point features, we use a 5-layer 3D full Conv network with the $32 \times 32 \times 32$ input grid resolution. The decoder is a 6-layer FC network with skip connections. Following~\cite{chibane2020neural}, we sample 5,000 training query points for each input image by Gaussian sampling with mixed variances in the vicinity of the ground truth surface. The maximal regressed distance is set to 0.2. The weights of $L_{AP}$ and $L_{GDA}$ are set to 1.0 and 0.02, respectively. We compute dense point clouds from the learned implicit surfaces using the dense point cloud extraction algorithm introduced in~\cite{chibane2020neural}. Please refer to the supplemental materials for more details about training and inference.

 \begin{figure}[t]
	\centering
	\includegraphics[width=8.3cm]{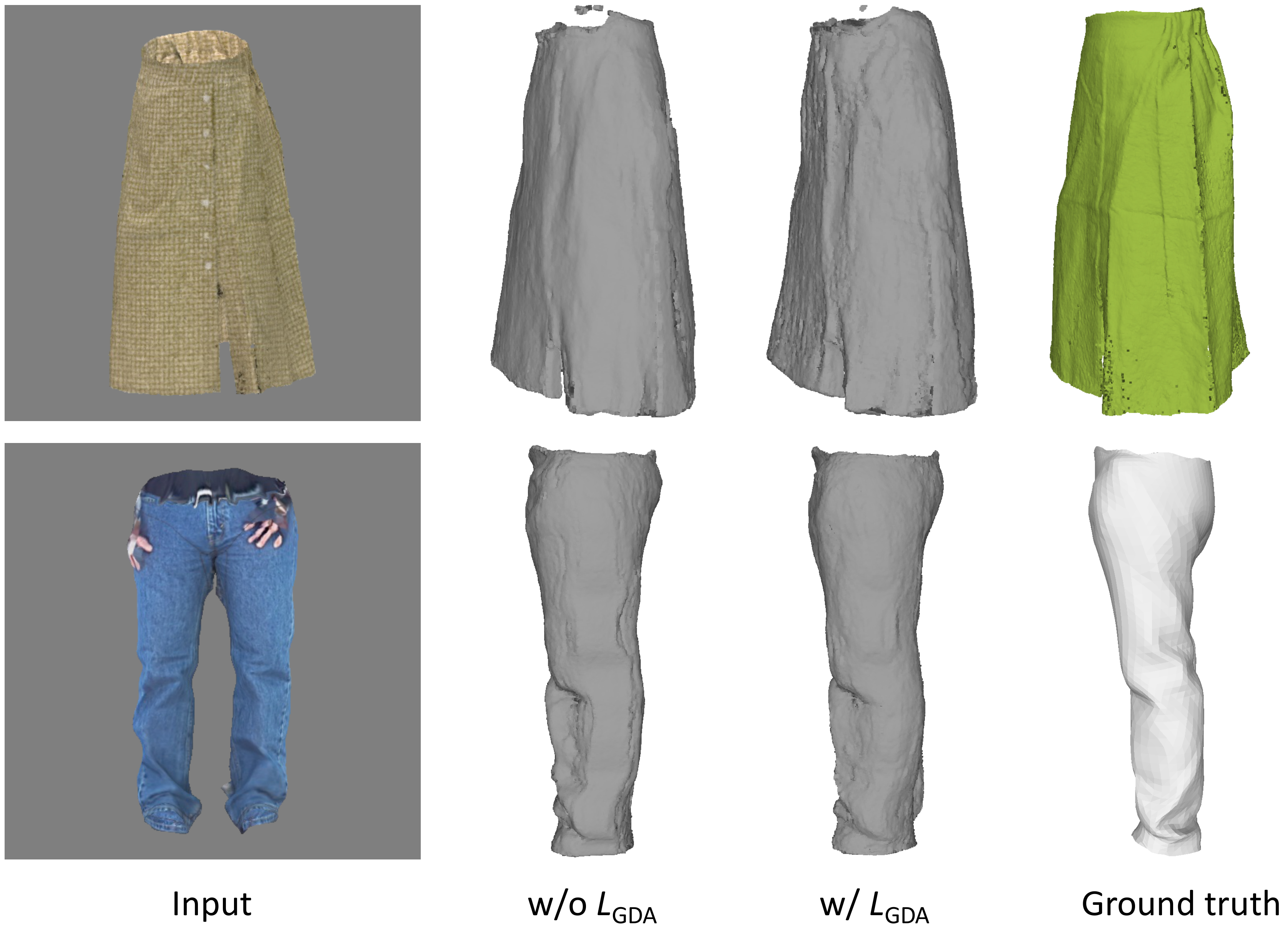}
	\caption{Visual comparison of with and without gradient direction alignment. By aligning the spatial gradient direction of the distance field, we can obtain sharper geometric details for the visible region and less artifacts for the self-occluded region.}\label{fig_comp_grad}
\end{figure}

\begin{table}[t]
	\centering
	\caption{Chamfer and P2S errors ($ \times {10^{ - 3}}$) with and without gradient direction alignment on MGN and Deep Fashion3D datasets.}
	\label{tab_results_gda}
	\begin{tabularx}{0.47\textwidth}{lYYYY@{}}
		\hlineB{2.5}
		\multirow{2}{*}{Methods} & \multicolumn{2}{c}{MGN} & \multicolumn{2}{c}{Deep Fashion3D} \\
		\cline{2-5}
		& Chamfer & P2S & Chamfer & P2S \\
		\hline
		w/o $L_{GDA}$ & 0.696 & 0.899 & 0.712 & 0.932 \\
		w/ $L_{GDA}$ & \textbf{0.635} & \textbf{0.762} & \textbf{0.621}  & \textbf{0.839} \\
		\hlineB{2.5}
	\end{tabularx}
\end{table}

\subsection{Ablation Study}

We validate the effectiveness of main components in our proposed AnchorUDF by both qualitative and quantitative evaluations. We first investigate the impact of the number of anchor points on MGN dataset. Table~\ref{tab_num_ap} lists Chamfer and P2S errors obtained by using 100, 300, 600 and 900 anchor points. It can be seen that using more anchor points achieves better results because stronger space context can be referenced by query points. However, when the point number exceeds a certain value, e.g., 900, the performance degrades. The reason is that too many points to regress increases the difficulty of learning the point prediction network and reduces its stability when testing, which further affects the subsequent shape reconstruction. Fig.~\ref{fig_ap_num} shows that compared to 100 anchor points, using 600 points produces finer shapes, especially for the side view.

 \begin{figure}[t]
	\centering
	\includegraphics[width=8.0cm]{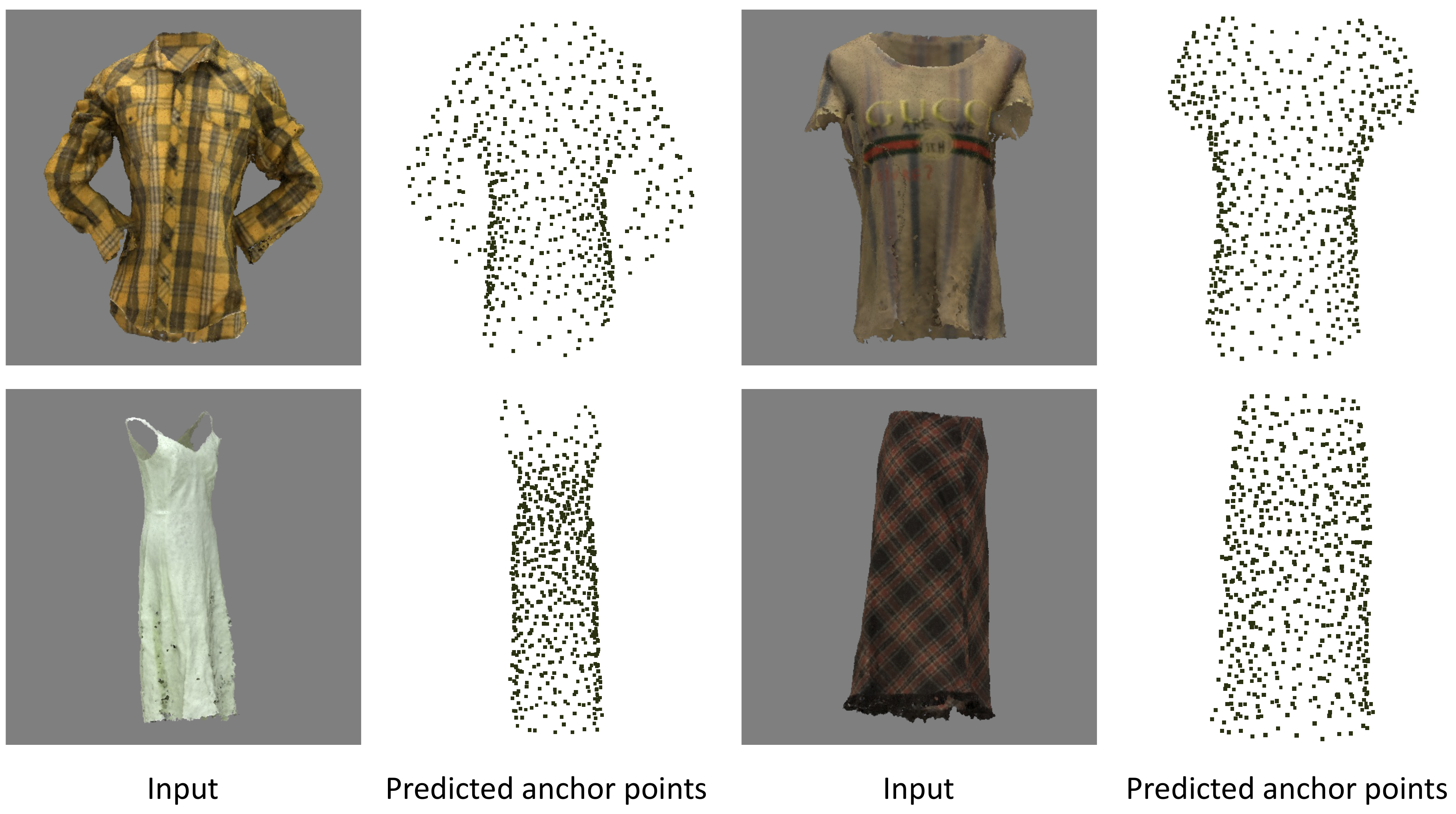}
	\caption{Visualization of anchor points predicted by our method.}\label{fig_ap_pred}
\end{figure}

We further assess the importance of the proposed anchored 3D position features by applying different 3D position features to our reconstruction framework, including the depth value, 3D point coordinates and our anchor point based position features. As reported in Table~\ref{tab_results_ap}, our position feature obtains the best results on both MGN and Deep Fashion3D. The depth value and 3D coordinates only provide absolute and local position information of query points, which is not enough to predict UDF with good neighborhood consistency. In contrast, anchor points located around the surface provide the information about shape profile which enables relative position computation between query points and the surface for more global and discriminative position features. Fig.~\ref{fig_pos_feat} illustrates some qualitative results. As one can see, the depth value fails to project some points which are far away from the surface, and 3D point coordinates perform a little better but still cannot obtain accurate surfaces at edges of the shapes.

To evaluate the influence of the proposed gradient direction alignment, Table~\ref{tab_results_gda} reports the results with and without $L_{GDA}$ in Eq.~\eqref{loss_grad}. It can be observed that optimizing $L_{GDA}$ further reduces Chamfer and P2S errors. Visual comparison is shown in Fig.~\ref{fig_comp_grad}. By explicitly aligning spatial gradient direction of the distance field during training, we can obtain sharper geometric details for the visible region and less artifacts for the self-occluded region.

We also visualize the anchor points predicted by our model to see if these points actually indicate shape profiles. As illustrated in Fig.~\ref{fig_ap_pred}, the predicted anchor points are evenly distributed around the input garments for different garment topologies and thus are able to provide holistic 3D shape information for query points.

 \begin{figure*}[t]
	\centering
	\includegraphics[width=17.0cm]{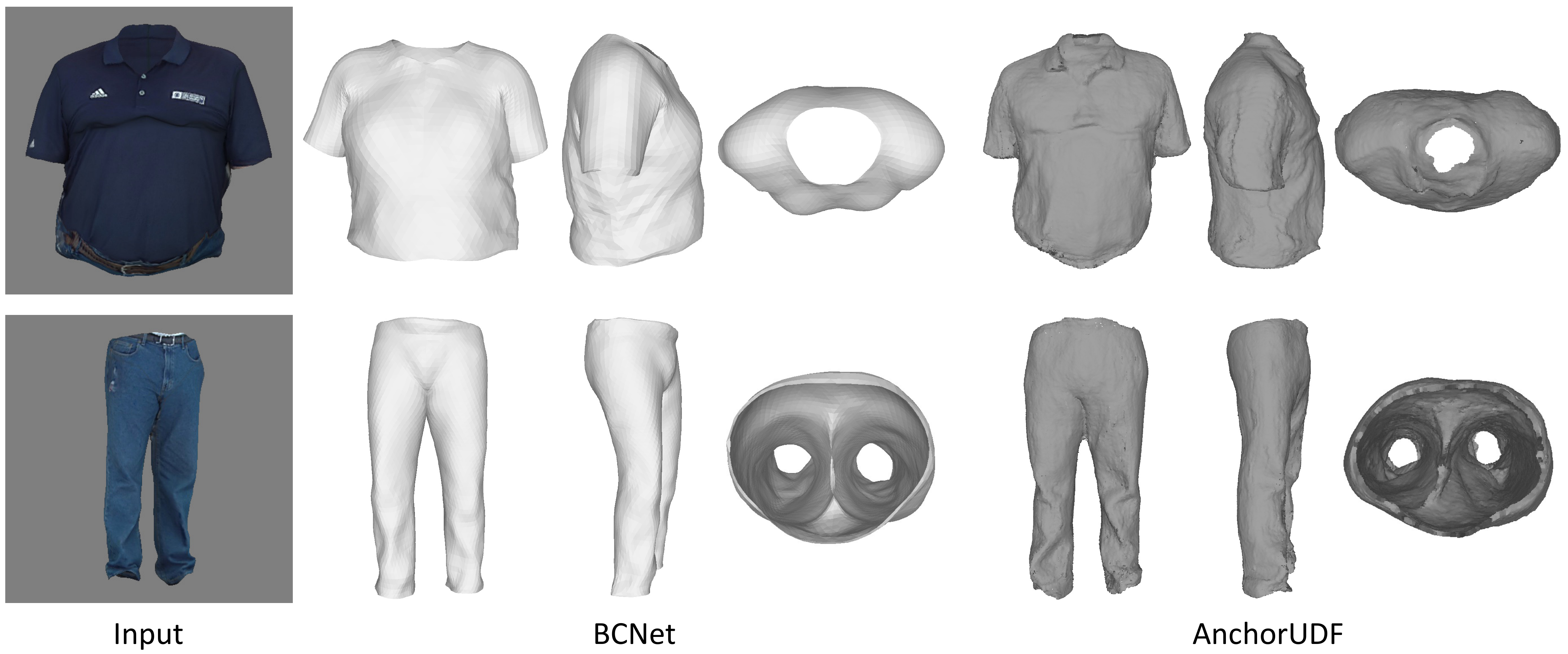}
	\caption{Visual comparison with BCNet~\cite{jiang2020bcnet} on MGN dataset. Although BCNet can recover garment shapes with open surfaces, it tends to produce overly smooth surfaces and loses lots of geometric details.}\label{fig_comp_mgn}
\end{figure*}

 \begin{figure*}[t]
	\centering
	\includegraphics[width=17.0cm]{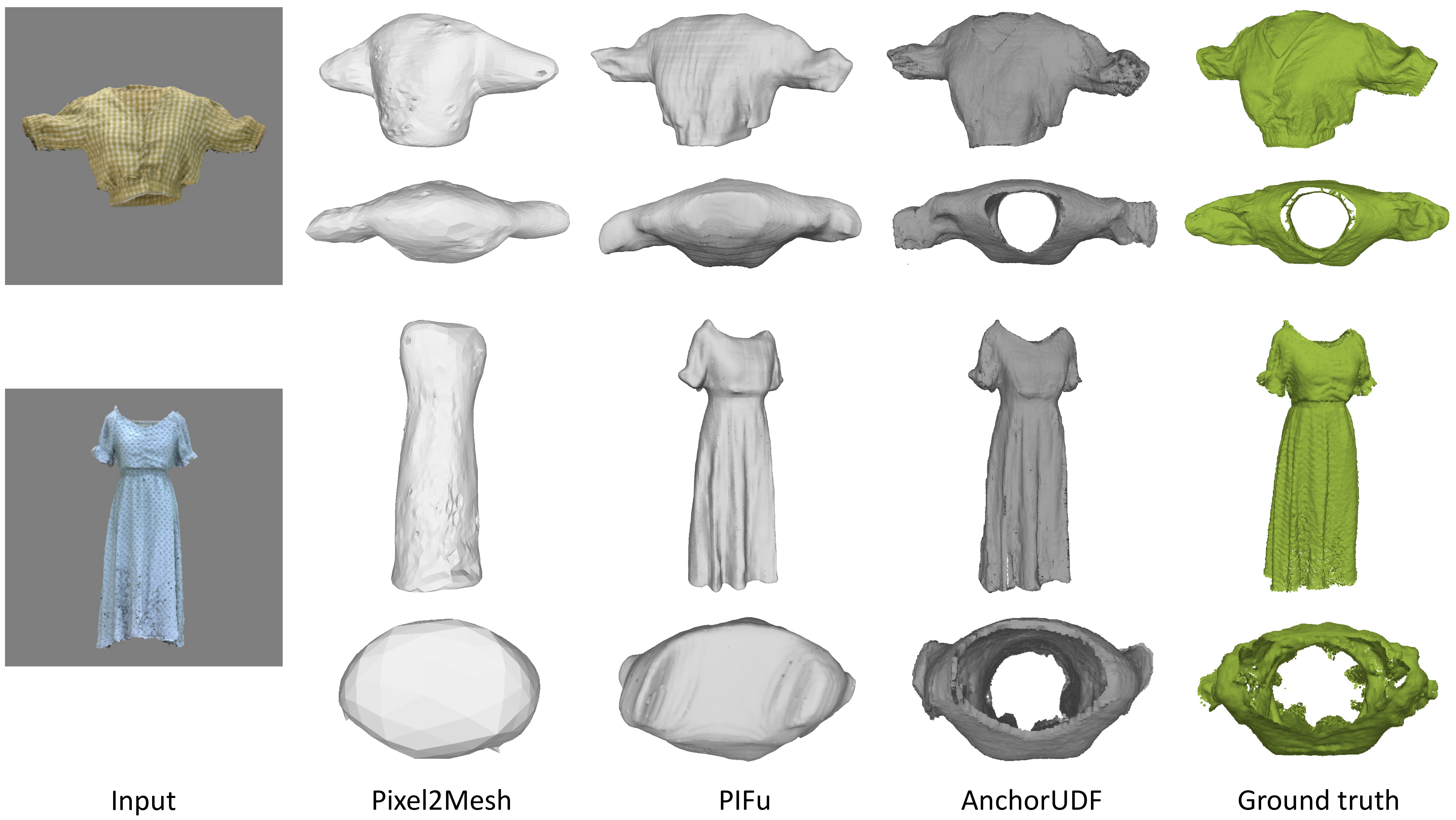}
	\caption{Visual comparison of different single-view reconstruction methods on Deep Fashion3D dataset. Neither Pixel2Mesh~\cite{wang2018pixel2mesh} nor PIFu~\cite{saito2019pifu} can handle garment reconstruction with open surfaces.}\label{fig_comp_df3d}
	\vspace{-8pt}
\end{figure*}

\begin{table}[t]
	\centering
	\caption{Chamfer and P2S errors ($ \times {10^{ - 3}}$) obtained by different single-view reconstruction methods on MGN and Deep Fashion3D datasets.}
	\label{tab_results_sota}
	\begin{tabularx}{0.35\textwidth}{lYY@{}}
		\hlineB{2.5}
		\multirow{2}{*}{Methods} & \multicolumn{2}{c}{MGN} \\
		\cline{2-3}
		& Chamfer & P2S  \\
		\hline
		BCNet~\cite{jiang2020bcnet} & 4.053 & 4.512 \\
		AnchorUDF & \textbf{0.635} & \textbf{0.762}  \\
		\hlineB{2.5}
		\multirow{2}{*}{Methods}& \multicolumn{2}{c}{Deep Fashion3D} \\
		\cline{2-3}
		& Chamfer & P2S  \\
		\hline
		Pixel2Mesh~\cite{wang2018pixel2mesh} & 4.266 & 5.330 \\
		PIFu~\cite{saito2019pifu} & 1.368 & 1.670 \\
		AnchorUDF & \textbf{0.621} & \textbf{0.839}  \\
		\hlineB{2.5}
	\end{tabularx}
\end{table}

\begin{figure*}[t]
	\centering
	\includegraphics[width=17.5cm]{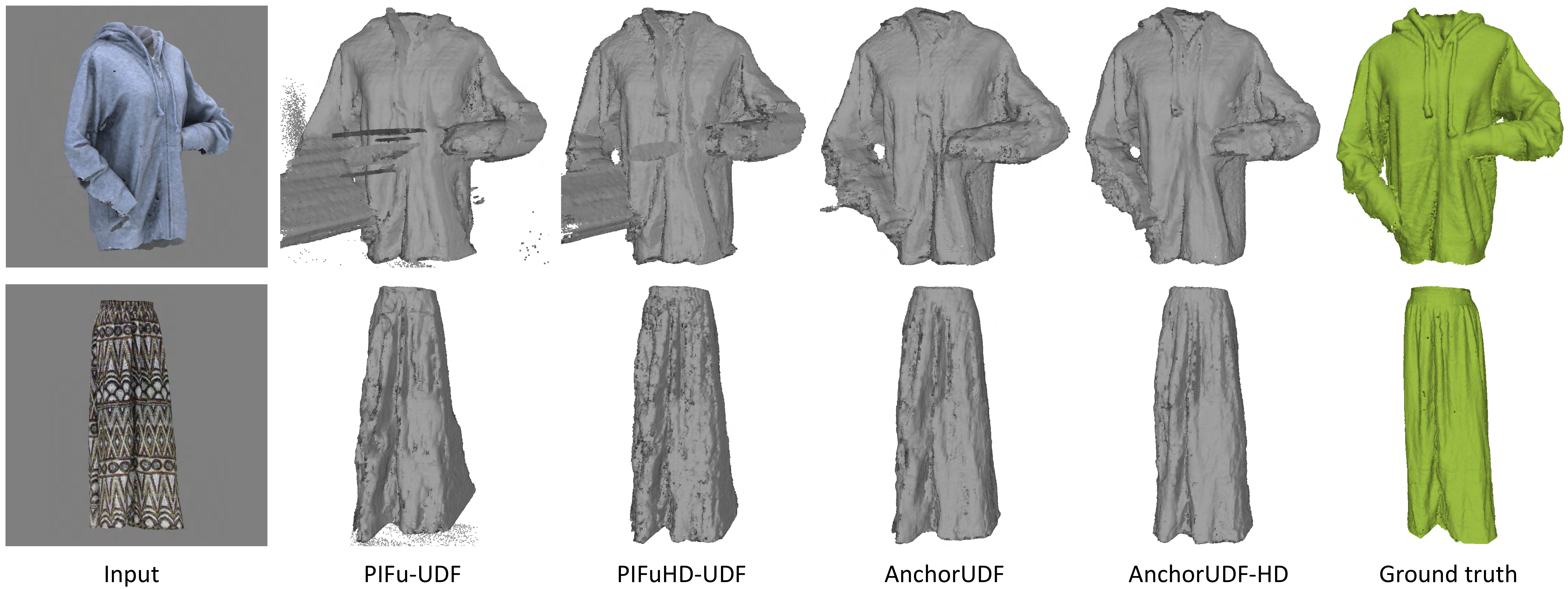}
	\caption{Visual comparison of methods extended with the HD or UDF module on Deep Fashion3D dataset. Our method can reconstruct more accurate shapes and richer details by adding the HD module to take high-resolution images as input.}\label{fig_hd_details}
\end{figure*}

\begin{table}[t]
	\centering
	\caption{Chamfer and P2S errors ($ \times {10^{ - 3}}$) obtained by methods extended with the HD or UDF module on Deep Fashion3D dataset.}
	\label{tab_comp_hd}
	\begin{tabularx}{0.4\textwidth}{lYY@{}}
		\hlineB{2.5}
		Methods & Chamfer & P2S \\
		\hline
		PIFu~\cite{saito2019pifu}-UDF &  1.411 & 2.440  \\
		PIFuHD~\cite{saito2020pifuhd}-UDF &  0.969  &  1.624  \\
		AnchorUDF &  0.621 & 0.839  \\
		AnchorUDF-HD &  \textbf{0.574} &  \textbf{0.737}  \\
		\hlineB{2.5}
	\end{tabularx}
\end{table}

\subsection{Comparison with Related Methods}

We compare our method against related single-view 3D reconstruction methods with different shape representations, including BCNet~\cite{jiang2020bcnet}, Pixel2Mesh~\cite{wang2018pixel2mesh}, PIFu~\cite{saito2019pifu} and PIFuHD~\cite{saito2020pifuhd}. As listed in Table~\ref{tab_results_sota}, our AnchorUDF obtains the lowest reconstruction errors on both MGN and Deep Fashion3D datasets. Fig.~\ref{fig_comp_mgn} qualitatively compares our method with BCNet~\cite{jiang2020bcnet} which represents garment shapes using template meshes on MGN dataset. Although such template-based method can recover garment shapes with open surfaces, it tends to produce overly smooth surfaces and loses lots of geometric details. Fig.~\ref{fig_comp_df3d} presents visual comparison on Deep Fshion3D dataset. Pixel2Mesh~\cite{wang2018pixel2mesh}, a mesh-based method whose shape is initialized from an ellipsoid, only recovers coarse shapes and cannot handle large deformations of garments. PIFu~\cite{saito2019pifu} using implicit function representation is able to generate detailed surfaces. However, it can only reconstruct closed surfaces, thus has difficulty handling garment reconstruction with multiple open boundaries. In contrast, our method can not only retain fine-scale geometric details but also faithfully capture open garment surfaces.

We further incorporate the HD module introduced in PIFuHD~\cite{saito2020pifuhd} into our AnchorUDF (AnchorUDF-HD) to evaluate the scalability of our method on high-resolution input images ($1024 \times 1024$). To make a fairer comparison, we try to directly use UDF as the implicit functions in PIFu (PIFu-UDF) and PIFuHD (PIFuHD-UDF) so that they can also represent open surfaces. Note that here we do not use extra normal maps as input because our main purpose is to verify if our method can be further improved under the HD framework. Table~\ref{tab_comp_hd} shows quantitative comparison on Deep Fashion3D. We can see that by taking high-resolution images as input, AnchorUDF-HD effectively reduces the reconstruction errors, especially P2S, indicating more accurate shape details are recovered. PIFuHD-UDF performs better than PIFu-UDF but still worse than our AnchorUDF, which shows that direct combination with UDF cannot work well and further confirms the necessity of the proposed components in AnchorUDF. Qualitative evaluation is shown in Fig.~\ref{fig_hd_details}. We can see that AnchorUDF-HD reconstructs more accurate shapes and richer details. Note that although PIFuHD-UDF uses 3D embedding provided by the coarse level as the input of distance function instead of the depth value, it sill cannot project points appropriately onto the surface. More qualitative results can be found in the supplemental materials.

\section{Conclusion}

We present Anchored Unsigned Distance Function (AnchorUDF) for single-view 3D garment reconstruction. For each query point, AnchorUDF not only extracts its pixel-aligned image features, but also computes its 3D position features based on a set of anchor points located around 3D surface to make the distance function better fit diverse garment shapes. Furthermore, we explicitly align the spatial gradient direction of AnchorUDF with the ground-truth direction to the surface during training to obtain more accurate projection directions for query points at inference. Experiments show that AnchorUDF achieves the state-of-the-art single-view reconstruction performance and is able to be scaled to high-resolution inputs.

{\small
\bibliographystyle{ieee_fullname}
\bibliography{zfbib_3d_iccv2021_arxiv}
}

\newpage

\section*{A. Model Training and Inference}
\label{model}

To generate training point sets, we randomly sample points on the ground-truth surface and displace them with Gaussian distribution $\mathcal{N}(0,\sigma )$ along x, y and z axis, where 1\% of points are sampled from $\sigma  = 0.08$, 49\% of points from $\sigma  = 0.02$ and 50\% of points from $\sigma  = 0.003$ as suggested by~\cite{chibane2020neural}. For training, we use the RMSprop optimizer with a learning rate of 5e-5. The batch size is 4 and the number of epochs is 35 for MGN and 60 for Deep Fashion3D. The learning rate is decayed by the factor of 0.1 in the last 20 epochs. We jointly optimize the UDF and anchor point regression losses from the beginning of training, and add the gradient direction loss to fine-tune the decoder in the last 10 epochs while fixing the encoders for training efficiency. All compared models are trained by the codes provided their authors. For PIFu~\cite{saito2019pifu} and PIFuHD~\cite{saito2020pifuhd}, we use the same backbone structure with our method. For BCNet~\cite{jiang2020bcnet}, we use the trained model provided by its authors because its training code is not released. At inference, the step number of projecting points is set to 5 and the valid distance to the surface is set to 0.007, which produce robust reconstruction results. Please refer to~\cite{chibane2020neural} for detailed algorithm steps of dense point cloud extraction.

\begin{figure}[t]
	\centering
	\includegraphics[width=8.2cm]{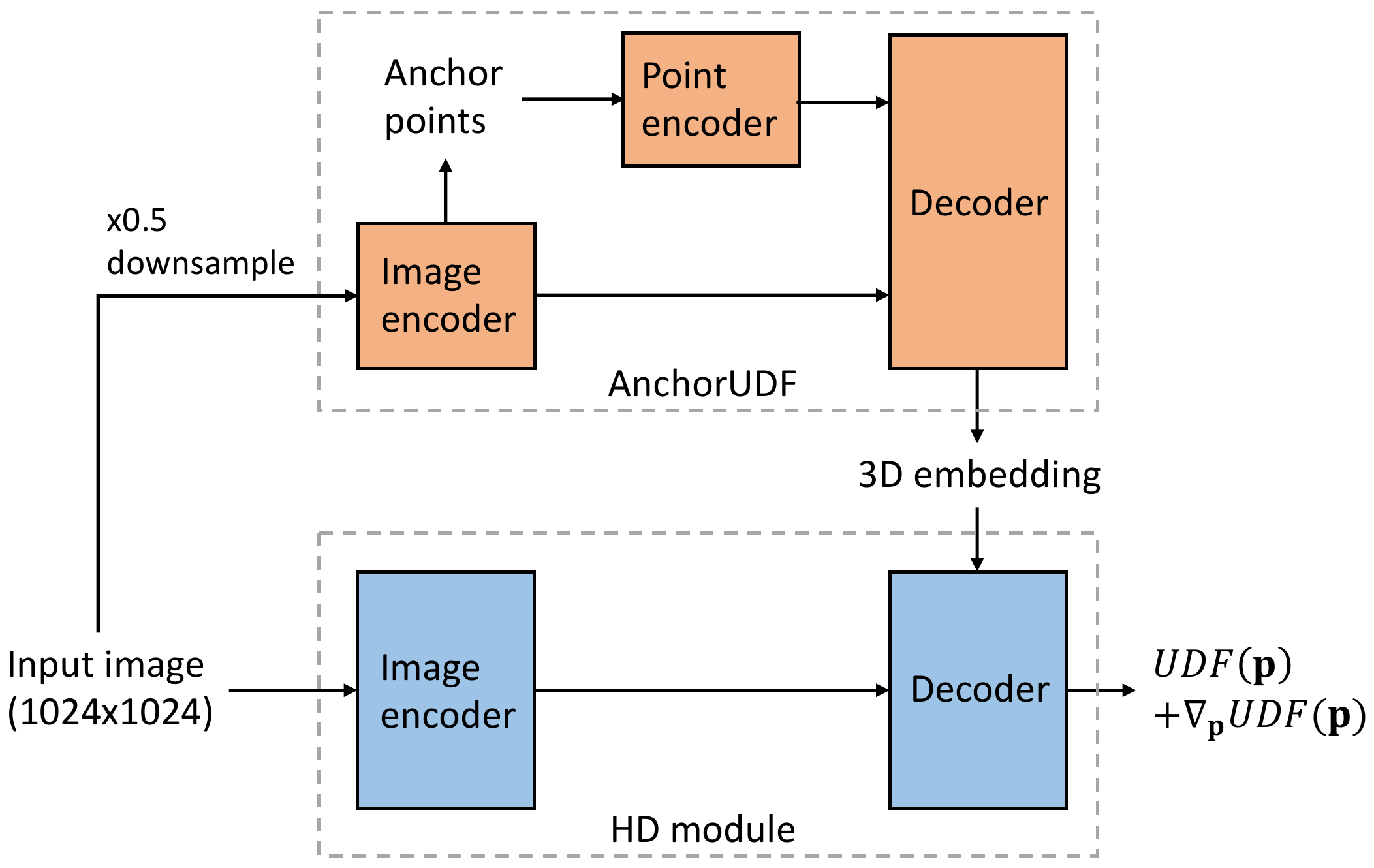}
	\caption{Flow chart of our AnchorUDF-HD. It incorporates the HD module into AnchorUDF, which takes high-resolution images as input.}\label{fig_flowhd}
\end{figure}

We illustrate the flow chart of our AnchorUDF-HD which incorporates the HD module~\cite{saito2020pifuhd} into AnchorUDF in Fig.~\ref{fig_flowhd}. Here a 3D embedding extracted from the decoder of AnchorUDF and local image features of high-resolution input ($1024 \times 1024$) are fed simultaneously into the decoder of the HD module to predict UDF. To learn AnchorUDF-HD, we first train AnchorUDF as the training procedure described before. Then, we add the HD module and continue to train the entire model for 15 epochs.

\section*{B. More Results}
\label{results}

\begin{figure*}[!htbp]
	\centering
	\includegraphics[width=13.5cm]{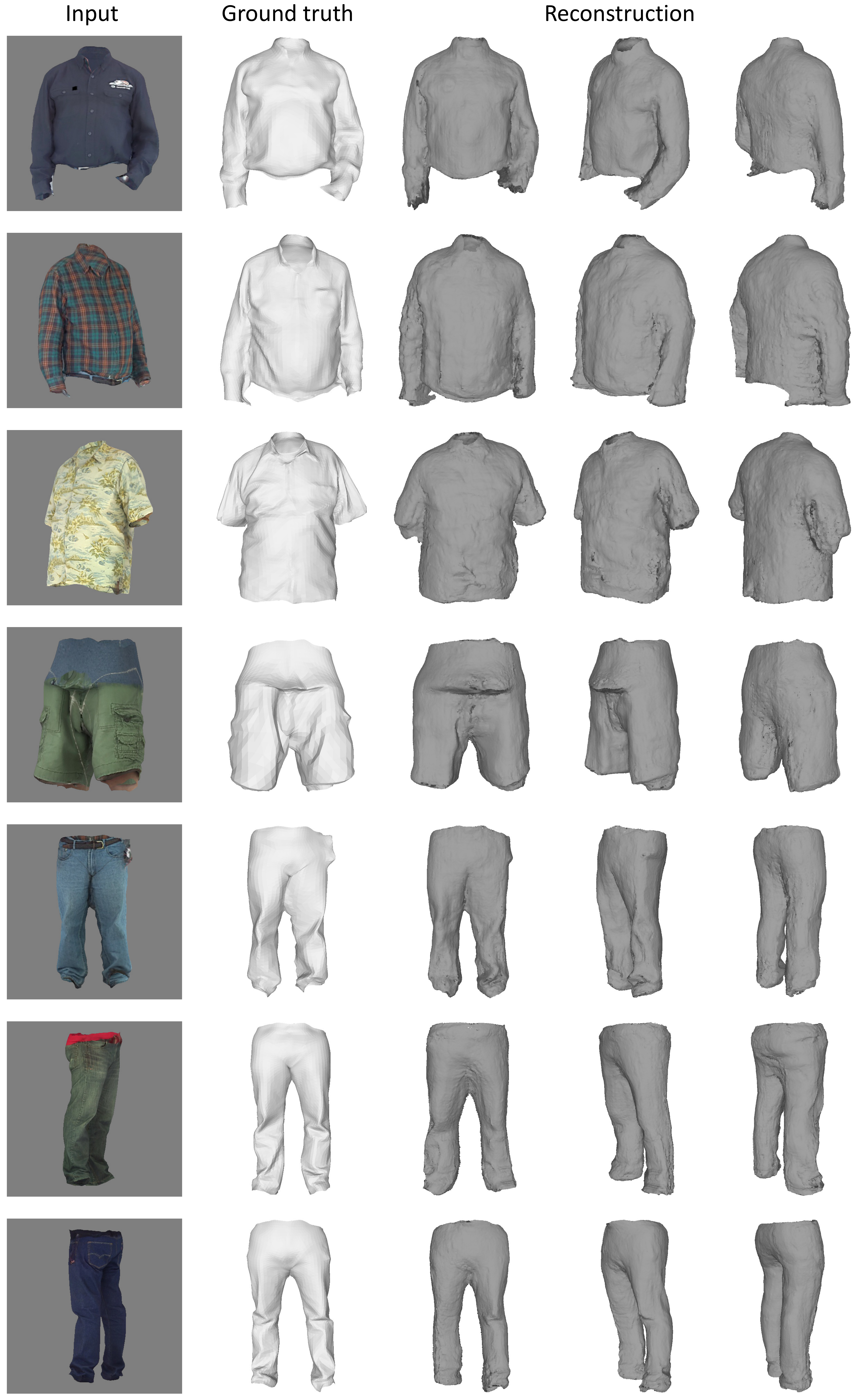}
	\caption{More garment reconstruction results on MGN dataset.}\label{fig_mgn}
\end{figure*}

\begin{figure*}[!htbp]
	\centering
	\includegraphics[width=16.0cm]{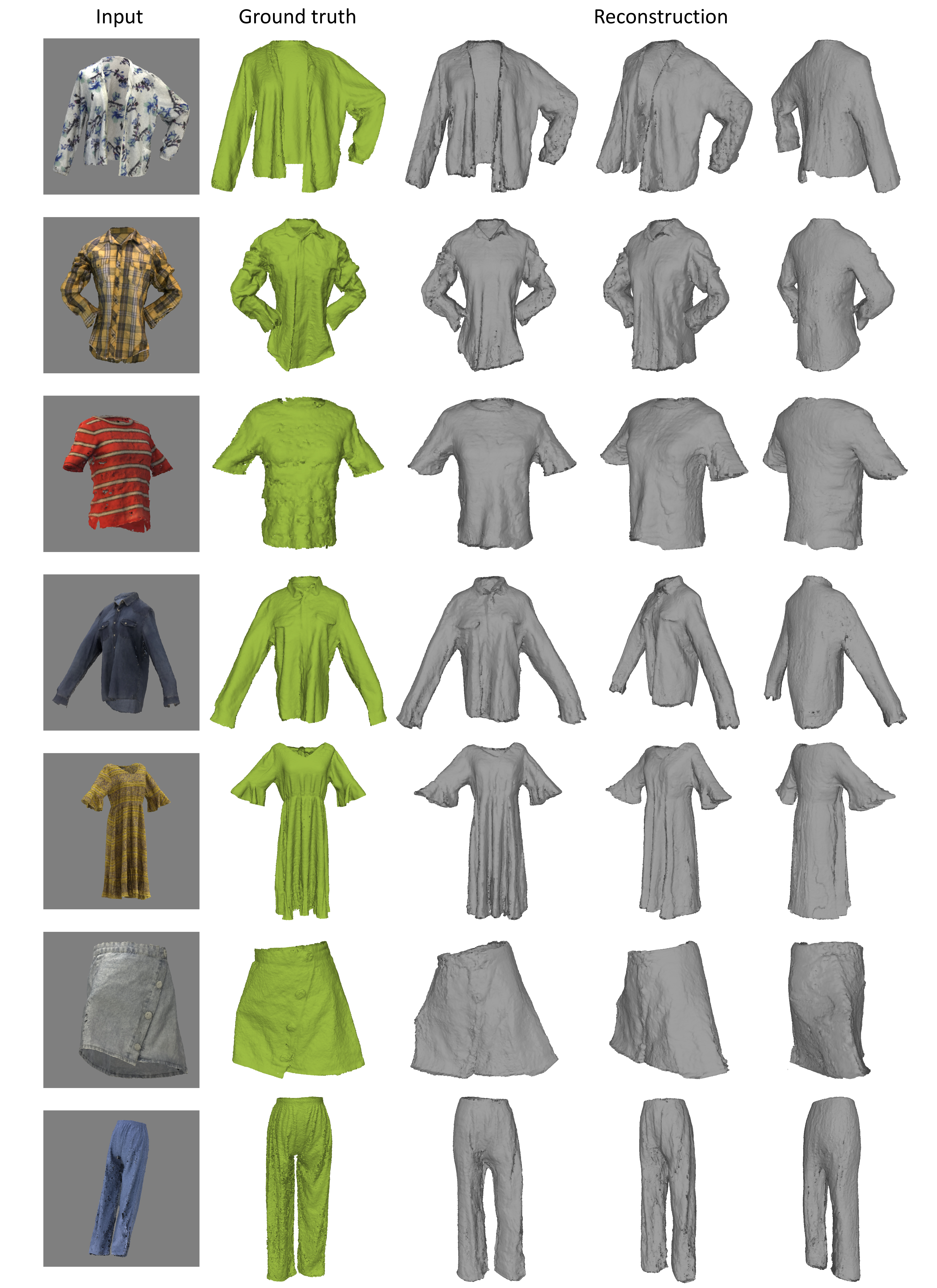}
	\caption{More garment reconstruction results on Deep Fashion3D dataset.}\label{fig_df3d}
\end{figure*}

We visualize more reconstruction results of our method on MGN (Fig.~\ref{fig_mgn}) and Deep Fashion3D (Fig.~\ref{fig_df3d}) datasets. As one can see, our method can faithfully recover detailed surfaces for inputs with various views and infer plausible shapes for self-occluded regions.

\begin{figure*}[!htbp]
	\centering
	\includegraphics[width=17.0cm]{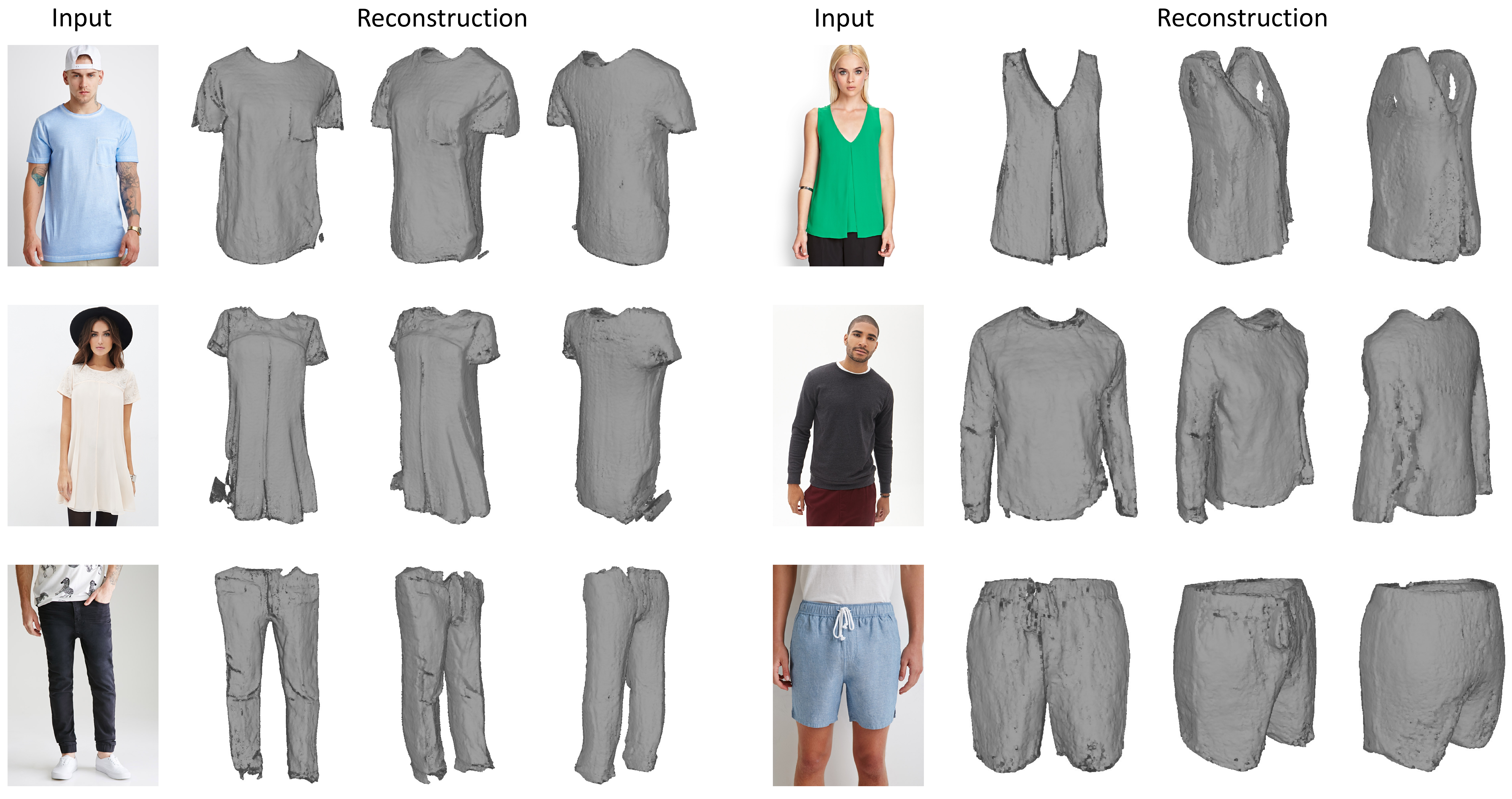}
	\caption{Reconstruction results on real garment images from DeepFashion dataset~\cite{liu2016deepfashion}. Our method can capture topologies of different garment categories and retain local details present in input images. Note that some artifacts are caused by hand occlusion from the original input images.}\label{fig_real}
\end{figure*}

We also test our model trained with Deep Fashion3D on real garment images from DeepFashion dataset~\cite{liu2016deepfashion}. Here we use semantic segmentation annotations provided by the dataset to obtain input garment images. As shown in Fig.~\ref{fig_real}, our method produces promising reconstruction results for different garment categories, which capture multiple topologies and retain local details present in input images. Note that we do not use real garment images during training and there are some noises in ground truth point clouds which affect the genuineness of rendered training images.

\end{document}